\documentclass[letterpaper, 10 pt, conference]{ieeeconf}  

\IEEEoverridecommandlockouts                              

\usepackage{graphicx}
\usepackage{cite}
\usepackage{amsmath,amssymb,amsfonts}
\usepackage[table]{xcolor}
\usepackage{threeparttable}

\title{\LARGE \bf
Open-Vocabulary Spatio-Temporal Scene Graph for Robot Perception and Teleoperation Planning
}

\author{Yi Wang$^{1}$, Zeyu Xue$^{1}$, Mujie Liu$^{1}$, Tongqin Zhang$^{1}$, \\ 
        Yan Hu$^{2}$, Zhou Zhao$^{3}$, Chenguang Yang$^{4}$ and Zhenyu Lu$^{1*}$
\thanks{\textsuperscript{1}HI-Robot Lab, School of Automation Science and Engineering, South China University of Technology.}%
\thanks{\textsuperscript{2}Institute of AI Industries, Chinese Academy of Science, 211135, China.}%
\thanks{\textsuperscript{3}School of Computer Science, Central China Normal University.}%
\thanks{\textsuperscript{4}School of Computer Science and Informatics, University of Liverpool.}
\thanks{*Corresponding author}
}

\begin{document}

\maketitle
\thispagestyle{empty}
\pagestyle{empty}

\begin{abstract}
Teleoperation via natural-language reduces operator workload and enhances safety in high-risk or remote settings. However, in dynamic remote scenes, transmission latency during bidirectional communication creates gaps between remote perceived states and operator intent, leading to command misunderstanding and incorrect execution. To mitigate this, we introduce the \textbf{Spatio-Temporal Open-Vocabulary Scene Graph (ST-OVSG)}, a representation that enriches open-vocabulary perception with temporal dynamics and lightweight latency annotations. ST-OVSG leverages LVLMs to construct open-vocabulary 3D object representations, and extends them into the temporal domain via Hungarian assignment with our temporal matching cost, yielding a unified spatio-temporal scene graph. A latency tag is embedded to enable LVLM planners to retrospectively query past scene states, thereby resolving local–remote state mismatches caused by transmission delays. To further reduce redundancy and highlight task-relevant cues, we propose a task-oriented subgraph filtering strategy that produces compact inputs for the planner. ST-OVSG generalizes to novel categories and enhances planning robustness against transmission latency without requiring fine-tuning. Experiments show that our method achieves 74\% node accuracy on Replica benchmark, outperforming ConceptGraph. Notably, in latency-robustness experiment, the LVLM planner assisted by ST-OVSG achieved a planning success rate of 70.5\%. 

\end{abstract}

\section{INTRODUCTION}

Teleoperation robots play a critical role in diverse application domains such as deep-sea exploration, nuclear plant accident response, and space station maintenance\cite{intro1}\cite{intro2}. By allowing humans to perform high-risk tasks remotely, they greatly enhance both safety and efficiency. As these applications expand, there is a growing demand for teleoperation systems to become more intelligent and autonomous, particularly with regard to their ability to perceive, interpret, and plan within complex environments. The emergence of large models offers new possibilities for these limitations.

 Recently, large-scale pretrained models, including Large Language Models (LLMs) and Vision-Language Models (VLMs), have demonstrated remarkable capabilities in perception and reasoning\cite{intro3}. Motivated by these advances, many researchers have integrated such models into teleoperation systems to improve generalization and robustness in open-world scenarios. These approaches have achieved promising results in object recognition, language understanding, and task planning. However, directly applying these models to teleoperation robotics still faces several challenges.

\begin{figure} [t]
     \centering
     \includegraphics[width=0.49\textwidth]{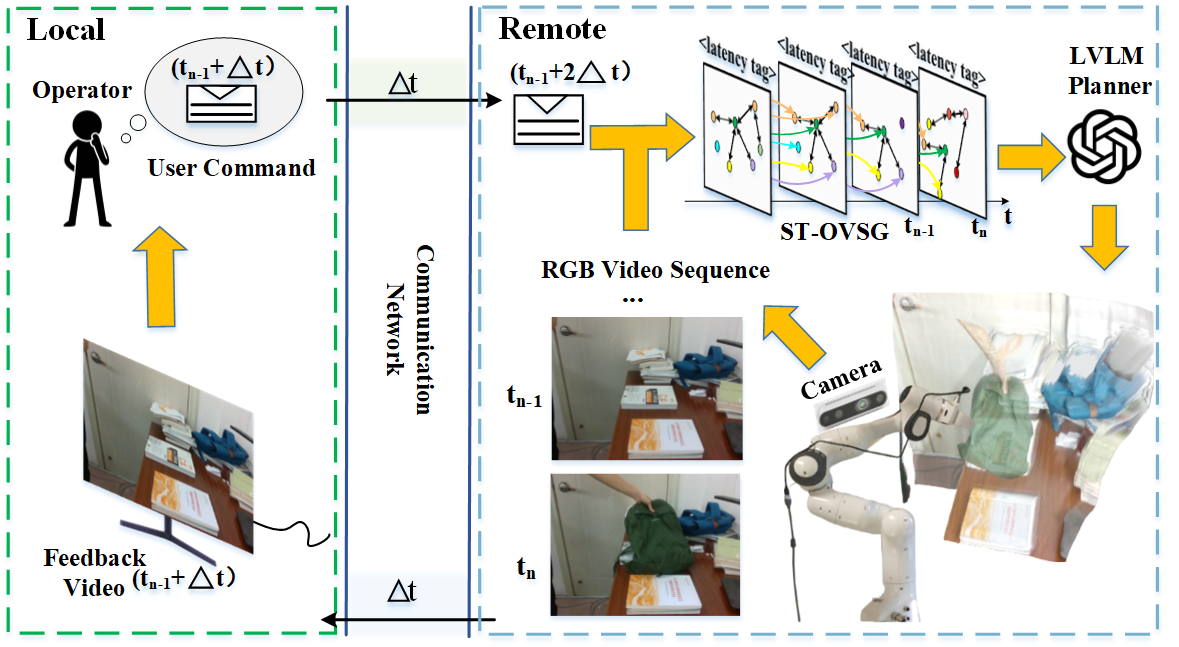}
     \caption{System overview. Based on the $t_{n-1}+\Delta{t}$ moment scene feedback, the local operator issues natural-language commands. These commands are sent over the data network to the remote side, where ST-OVSG temporally aligns the local commands with the remote observations to compensate for link latency. This alignment stabilizes the large model’s semantic reasoning and drives reliable execution by the robotic arm.}
     \label{fig1}
     \vspace{-1.5em}
\end{figure}

The first challenge is communication latency. In real-world networked environments, teleoperation systems are inevitably affected by delays. Such delays create a mismatch between operator-issued commands and the robot’s perception of the evolving environment \cite{intro4}, directly reducing task efficiency and accuracy. Existing methods often assume that user commands are received and executed immediately after being issued, which does not hold under real teleoperation network conditions. 

For example, during a deep-sea exploration mission, an operator may instruct the robot to “collect the rock sample lying on the seabed.” Due to long-distance communication latency, however, by the time the command reaches the robot, ocean currents may have displaced it to a nearby but different area. Without the ability to reason about latency and temporal order, the robot may mistakenly collect the wrong rock. In high-stakes scenarios such as deep-sea or nuclear operations, the cost of such errors is critical, as tolerance for mistakes is extremely low.

The second challenge is the static nature of current scene representations. Most existing methods rely on object positions, visual features, and text labels \cite{nlmap,vlp,conceptgraphs}. While effective in static environments, these approaches lack the capacity to capture continuous events and temporal dynamics. They do not track how objects or interactions evolve over time, which is essential for teleoperation in dynamic settings. Moreover, raw video streams contain substantial redundancy, particularly in long-duration tasks, which dilutes key events and makes it harder for the system to focus on task-relevant cues. By structuring observations into scene graphs, salient objects, relations, and events can be highlighted, providing a more concise and informative representation.

Taken together, these challenges reveal a fundamental gap: latency distorts the temporal alignment between operator intent and robot execution, while static representations fail to capture evolving events or filter redundant information. To address this, we propose \textbf{Spatio-Temporal Open-Vocabulary Scene Graph (ST-OVSG)}, an open-vocabulary spatio-temporal scene graph designed for teleoperation.

The main contributions of this work can be summarized as follows:

\begin{enumerate}
    \item We propose \textbf{ST-OVSG}, a novel spatio-temporal open-vocabulary scene graph, which explicitly models both the spatial structure and temporal dynamics of scenes, capturing how objects and events evolve over time.
    \item To improve the planner’s awareness of communication delays, we introduce a lightweight latency tag into the scene graph. This design reduces the negative impact of delayed teleoperation commands and enables more robust decision making under real-world network conditions.
    \item Extensive experiments demonstrate that ST-OVSG effectively models temporal variations and strengthens delay awareness in robotic planning, while achieving performance comparable to state-of-the-art methods in other aspects.
\end{enumerate}

\section{Related Work}

\begin{figure*} [ht!]
     \centering
     \includegraphics[width=\textwidth]{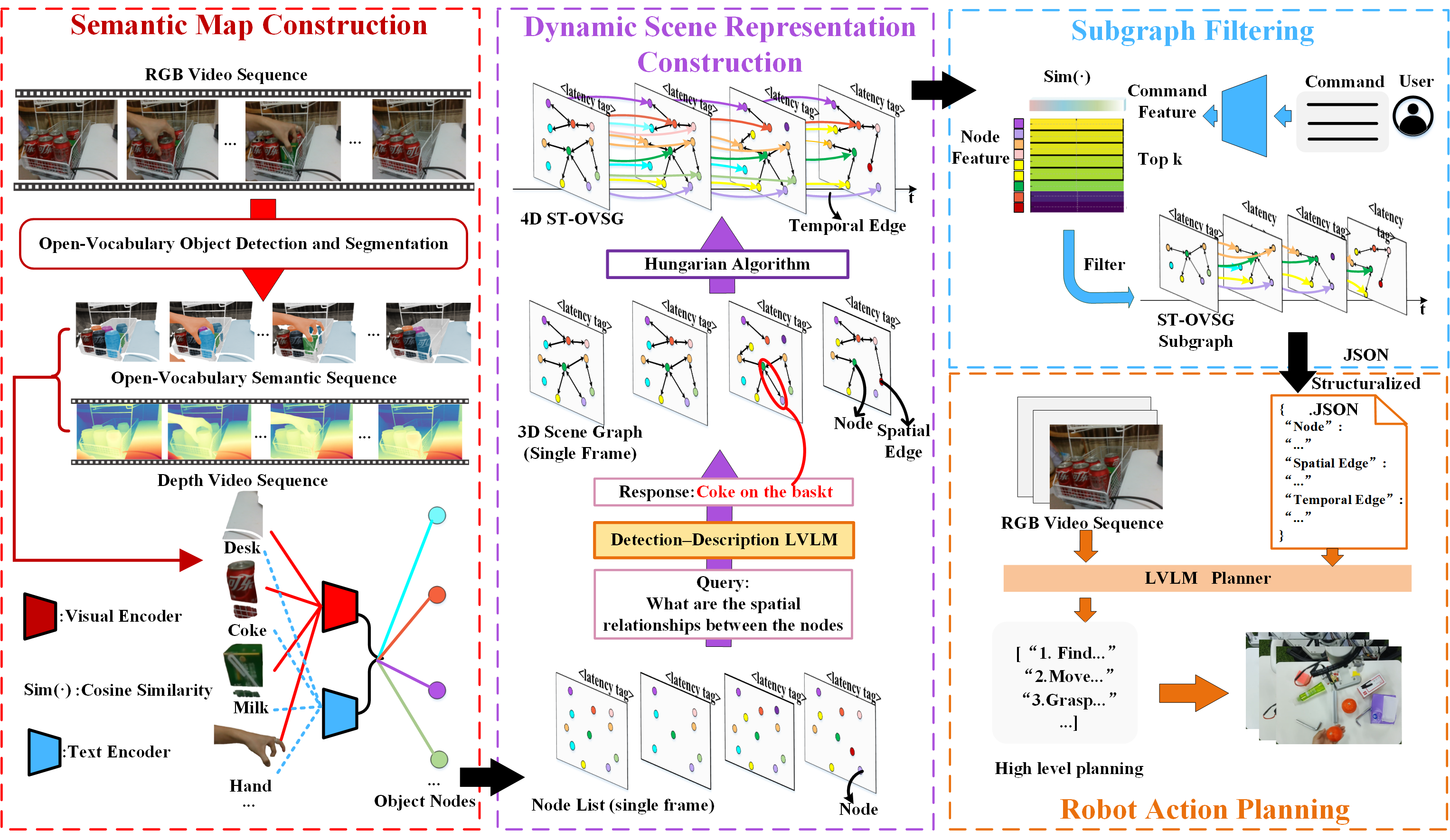}
     \caption{ST-OVSG builds a spatio-temporal open-vocabulary scene graph from RGB-D video sequences. Objects are detected and segmented from RGB frames, fused with depth to form semantic nodes. These graphs are linked across frames using the Hungarian algorithm\cite{Hungarian1}\cite{Hungarian2}, producing a 4D scene graph with spatial and temporal edges and latency tags. User commands are used to query node features, filtering relevant nodes to form an ST-OVSG subgraph, which is then serialized into JSON and provided to the LVLM planner for generating executable robot task plans.}
     \label{fig2}
     \vspace{-1.5em}
\end{figure*}

\subsection{Robot Planning with Foundation Models}

In robotics, traditional methods are often restricted to simple, structured scenarios. Deep learning approaches achieve better performance but rely heavily on large datasets and struggle with generalization. Recently, foundation models have demonstrated strong reasoning and generalization capabilities\cite{brown} \cite{liu2023reflect}, and their success in Natural Language Processing (NLP) and Computer Vision (CV) has spurred growing interest in applying them to robotics. Language models are capable of handling complex semantic understanding and task reasoning with strong generalization. It provide an opportunity to address many of the shortcomings found in traditional and deep learning methods. Therefore, more and more researchers are exploring how to apply foundation models (FMs) in the field of robotics

SayCan \cite{saycan} couples an LLM for high-level reasoning with value functions over pretrained skills for feasibility, grounding language into actions on a real mobile manipulator for long-horizon instructions. PaLM-E \cite{PaLM-E} introduces an embodied multimodal language model that fuses perceptual inputs such as vision with the PaLM LLM. This integration enables end-to-end reasoning from raw sensory data to actions, achieving strong cross-modal generalization and robust performance across diverse robotic tasks. ProgPrompt \cite{progprompt} uses program-like prompts that specify available actions and objects, mitigating free-form hallucinations and producing executable plans; it is validated in VirtualHome and on a physical arm. 

While these approaches significantly enhance the reasoning and generalization capabilities of robotic systems, they all overlook the impact of network latency on task execution. This limitation is particularly critical in teleoperation scenarios, where delays can cause a mismatch between dynamic environmental events and operator intent.

\subsection{Semantic Scene Representations} 
Semantic scene representation is central to computer vision and AI, and increasingly vital for robot perception and planning. Traditional Simultaneous Localization and Mapping (SLAM)  \cite{slam1,slam2,slam3,slam4}builds geometric maps for navigation but lacks semantic differentiation, treating objects such as chairs and tables merely as obstacles. With advances in deep learning, semantic maps integrate object detection or segmentation results\cite{detection1,detection2,detection3,detection4}, enriching maps with bounding boxes and labels\cite{boundingbox1}\cite{boundingbox2}. More recently, scene graphs\cite{scene_graph1}\cite{scene_graph2} represent objects as nodes and their relationships as edges, offering structured understanding of environments. However, both remain limited by closed vocabularies, constraining their generalization to open and dynamic real-world scenarios.
\subsection{Scene Representations in Robotics}

Recent advances in foundation models, such as CLIP and ViLD\cite{clip}\cite{vild}, have enabled open-vocabulary detection, allowing semantic scene representations to move beyond predefined categories and recognize unseen objects. This progress has motivated applications in robotics. For instance, Chen et al.\cite{nlmap} proposed queryable scene representations that integrate visual features, textual labels, and 3D positions, and employed large language models to decouple user instructions for planning. Huang et al.\cite{vlp} introduced visual-language maps that combine semantic features with spatial coordinates, enabling language-driven navigation. Gu et al.\cite{conceptgraphs} further developed open-vocabulary 3D scene graphs that incorporate object relationships, yielding richer structures for perception and planning.

These works significantly improve the generalization of robotic perception and planning. However, they remain limited to static environments. Objects are encoded at build time, and once they move or disappear, the representation becomes outdated. This restricts performance in dynamic or delayed teleoperation settings.
\section{Methodology}

\subsection{Problem Formulation}
We aim to construct a temporally indexed, semantically enriched representation of dynamic 3D environments, enabling LVLM-based robot planner to plan action based on remote commands robustly, even under non-negligible communication delays. In teleoperation settings, the LVLM planner operates remotely together with the robot, while a human operator issues command from a local site. Delays on the order of hundreds of milliseconds to several seconds may arise during the bidirectional transmission of visual data and operator commands. Meanwhile, the remote environment can evolve significantly, creating a mismatch between the state visible to the operator when issuing a command and the state available to the robot when receiving it.

Formally, the challenge is to maintain a representation that allows the system to (i) recover the scene as it existed at the command-issue time, (ii) resolve temporal ambiguities among similar or evolving objects, and (iii) provide a reliable substrate for long-horizon planning. Consider a case where two visually similar objects (e.g., two red mugs) appear at different times: without temporal grounding, the intended referent cannot be resolved once latency is introduced. To address this, we propose ST-OVSG that integrates object nodes, spatial relations, and temporal correspondences. This representation supports locating arbitrary objects across time, grounding commands in a latency-aware manner, and enabling planning in dynamic environments. The approach is illustrated in Fig. \ref{fig2}.

\subsection{Spatio-Temporal Scene Representation}
Given a time-ordered set of posed RGB-D frames \(\mathcal{D}=\{(I_n^{\mathrm{rgb}}, I_n^{\mathrm{d}}, \Delta t_n, \tau_n)\}_{n=1}^{N}\) where \(\Delta T_n\) and \(\tau_n\) is the transmission latency and capture timestamp at remote robot respectively. This set acquired from a calibrated, RGB-D robot camera with known pose observing a dynamic environment. We maintain \emph{ST-OVSG}, the representation is \(\mathcal{G}_{1:N}=(\{\mathcal{M}_n\}_{n=1}^{N}, \mathcal{E}^{\mathrm{temp}})\), where each per-frame graph is \(\mathcal{M}_n=(\mathcal{O}_n, \mathcal{E}^{\mathrm{spa}}_n)\). Here, \(\mathcal{O}_n=\{\mathbf{o}_{i,n}\}_{i=1}^{N_o}\), \(N_o\) is number of all nodes of a frame, denotes object nodes at frame \(n\), \(\mathcal{E}^{\mathrm{spa}}_n \subseteq \mathcal{O}_n \times \mathcal{O}_n \times \mathcal{R}^{\mathrm{spa}}\) encodes spatial relations (with open-vocabulary labels, e.g., \textit{on}, \textit{inside}, \textit{left-of}), and \(\mathcal{E}^{\mathrm{temp}} \subseteq \bigcup_{n=1}^{N-1}\mathcal{O}_n \times \mathcal{O}_{n+1} \times \mathcal{R}^{\mathrm{temp}}\) links instances over time (e.g., \textit{same-instance}, \textit{appeared}, \textit{disappeared}). \(\mathcal{G}_{1:N}\) is constructed incrementally, the resulting time-stamped, open-vocabulary scene graph supports object retrieval across time, latency-aware instruction grounding, and long-horizon planning in dynamic scenes.

\subsubsection{Object Representation}
Each object node \(\mathbf{o}_{i,n}\) carries open-vocabulary semantics and grounded 3D geometry. Given an RGB-D frame \((I_n^{\mathrm{rgb}}, I_n^{\mathrm{d}})\), a detection–description LVLM \(\Psi_{\mathrm{desc}}\) predicts an open-vocabulary class label \(y_{i,n}\) for each candidate object and, when applicable, pairwise spatial relations \(r_{j,k,n}\in\mathcal{R}^{\mathrm{spa}}\) between objects \(j\) and \(k\). An object detection and segmentation model proposes regions \(\{(\mathbf{b}_{i,n}, \mathbf{m}_{i,n})\}_{i=1}^{N_o}\), where \(\mathbf{b}_{i,n}\) denotes a region of interest (ROI) and \(\mathbf{m}_{i,n}\) the corresponding mask. An image encoder \(\Phi_{\mathrm{v}}\) and a text encoder \(\Phi_{\mathrm{t}}\) (e.g., CLIP \cite{clip}) are adopted to extract masked visual features \(\mathbf{f}^{\mathrm{img}}_{i}=\Phi_{\mathrm{v}}(I_n^{\mathrm{rgb}};\mathbf{b}_{i,n})\) and natural-language features \(\mathbf{f}^{\mathrm{txt}}_{i}=\Phi_{\mathrm{t}}(y_{i,n})\).

To obtain metric 3D geometry, let \(\mathbf{u}=(u,v)^\top\) be the pixel coordinate and \(d_n(\mathbf{u})\) its corresponding depth value. Pixels under \(\mathbf{m}_{i,n}\) are lifted into the world frame using the camera intrinsics \(\mathbf{K}\), and the camera-frame point is then computed as:
\begin{equation}
\mathbf{x}_c = d_n(\mathbf{u})\,\mathbf{K}^{-1}[u, v, 1]^\top.
\end{equation}
This process yields an object-oriented semantic point set \(\mathcal{P}_{i,n}\), from which the centroid \(\mathbf{c}_{i,n}\) and size \(\mathbf{s}_{i,n}\) are estimated. In addition, to support cross-time object localization and historical scene reconstruction under teleoperation delays, each node stores an observation timestamp \(\mathbf{obs}_{n} = \tau_n + \Delta t_n\). Thus, \(\mathbf{obs}_{n}\) represents the effective time at which the corresponding visual data becomes available to the local human operator. This allows the planner to interpret user-issued commands with respect to the scene state observed by the operator.

The final object node representation is summarized as
\(
\mathbf{o}_{i,n} = \{\mathbf{b}_{i,n}, \mathbf{m}_{i,n}, \mathbf{f}^{\mathrm{img}}_{i}, \mathbf{f}^{\mathrm{txt}}_{i}, \mathbf{c}_{i,n}, \mathbf{s}_{i,n}, \mathcal{P}_{i,n}, \mathbf{obs}_{n}\},
\)
which cleanly factors visual features, language features, metric 3D geometry, and temporal alignment information to support robust spatio-temporal reasoning in teleoperation scenarios.

\subsubsection{Object Association and Correspondence}
We model both spatial relations within a frame and temporal correspondences across frames to obtain a coherent 4D graph. For spatial relations, the LVLM \(\Psi_{\mathrm{desc}}\) proposes candidate relation labels for object pairs \((j,k)\in\mathcal{O}_n\) together with an interaction bounding box \(\mathbf{b}^{\mathrm{int}}_{j,k,n}\), and matching is performed based on bounding-box overlap. However, when multiple candidate pairs overlap or are ambiguous, resolve them by minimizing a simple geometric cost. Let the union box be \(\mathbf{u}_{j,k,n}=\mathrm{Union}(\mathbf{b}_{j,n},\mathbf{b}_{k,n})\) and denote the interaction zone by \(\mathbf{z}_{j,k,n}=\mathbf{b}^{\mathrm{int}}_{j,k,n}\). The cost is


\begin{equation}
\begin{aligned}
c^{\mathrm{spa}}_{j,k,n} 
&= w_{\mathrm{iou}}\left(1-\mathrm{IoU}(\mathbf{u}_{j,k,n},\mathbf{z}_{j,k,n})\right) \\
&\quad + w_{\mathrm{area}}\left|\log \frac{A(\mathbf{u}_{j,k,n})}{A(\mathbf{z}_{j,k,n})}\right| \\
&\quad + w_{\mathrm{ctr}}\frac{\left\|\mathrm{ctr}(\mathbf{u}_{j,k,n})
-\mathrm{ctr}(\mathbf{z}_{j,k,n})\right\|}{\mathrm{diag}(\mathbf{z}_{j,k,n})},
\end{aligned}
\end{equation}
where \(A(\cdot)\) is area, \(\mathrm{ctr}(\cdot)\) is the box center, and \(\mathrm{diag}(\cdot)\) is the diagonal length. For any \emph{ambiguous} candidate set that shares a zone or an object, the unique partner with the minimum cost is kept and its spatial relation is retained as spatial edge.


For temporal correspondence, an active track set \(\mathcal{T}_{n-1}=\{\iota\}\) is maintained. Each track stores the attributes of the object's most recent node together with its last-seen timestamp. A cost matrix \(C\) is constructed with terms for visual-feature similarity, class consistency, and 3D distance between the current detections and the active tracks:


\begin{equation}
\begin{aligned}
C_{\iota,i} 
&= w_{\mathrm{pos}}\;\min\!\Biggl(
    \frac{\|\hat{\mathbf{c}}_{\iota}^{-}-\mathbf{c}_{i,n}\|_2}{d_{\max}},\,1
\Biggr) \\
&\quad + w_{\mathrm{vis}}\;\Bigl(1-\cos(\hat{\mathbf{f}}^{\mathrm{img}}_{\iota},\,\mathbf{f}^{\mathrm{img}}_{i,n})\Bigr) \\
&\quad + \delta_{\mathrm{cls}}\;\mathbb{1}[\,y_{\iota}\neq y_{i,n}\,],
\end{aligned}
\end{equation}
where \(d_{\max}\) normalizes 3D distance, \(\cos(\mathbf{a},\mathbf{b})\) is cosine similarity, \(w_{\mathrm{pos}},w_{\mathrm{vis}}\) are weights for each term, and \(\delta_{\mathrm{cls}}\) is a finite penalty that allows future class correction rather than hard rejection. The assignment is solved with the Hungarian algorithm to obtain candidate matchings \(\mathcal{A}_n\subseteq \mathcal{T}_{n-1}\times\mathcal{I}_n\). 

A matching \((\iota,i)\in\mathcal{A}_n\) is accepted if \(C_{\iota,i}<\eta\), where \(\eta\) is a threshold. Accepted matches update the corresponding tracks and refresh descriptors. Detections not in any accepted pair form the unmatched set \(\mathcal{U}_n\) and spawn new tracks with fresh identifiers. Tracks not in any accepted pair are marked as \emph{disappeared}, yielding the set \(\mathcal{L}_n\). The procedure thus returns the triplet (accepted matches, unmatched detections, disappeared tracks) \(=(\mathcal{A}_n^{\star},\mathcal{U}_n,\mathcal{L}_n)\). When an object reappears within the grace period, it can be matched to a previously disappeared track, creating a temporal edge across frames.

To explicitly handle teleoperation scenarios, each frame-level graph \(\mathcal{M}_n\) additionally stores its capture timestamp and estimated transmission latency \((\tau_n,\Delta T_n)\) as $<$latency tag$>$. Here \(\tau_n\) corresponds to the remote capture time at the robot side, and \(\tau_n+\Delta T_n\) reflects the time at which the visual frame is observed by the local operator. This latency tag allows the planner to align user-issued commands with the correct historical scene state available to the operator, thereby reducing referential errors in delayed execution and stabilizing object association in dynamic environments.

\subsubsection{Robot Planning through LVLMs}
To enable execution of user commands under teleoperation, the ST-OVSG is interfaced with an LVLM planner that resides together with the robot on the remote side. Let teleoperator at the local site issue a command \(u\) at time \(\tau_u\), which is transmitted to the remote side with an end-to-end latency \(\Delta T_u\). During this round trip, the remote environment may evolve, making it essential to align the planner’s reasoning with the scene state that was visible to the operator rather than the state at the time of receipt.

Because serializing the entire graph \(\mathcal{G}_{1:N}\) into text is costly, a compact, task-oriented subgraph centered on relevant objects is extracted. The command is embedded with the CLIP text encoder, \(\mathbf{g}_u=\Phi_{\mathrm{t}}(u)\), and object nodes are scored by
\begin{equation}
s_{i,t}=\cos(\mathbf{g}_u,\mathbf{f}^{\mathrm{txt}}_{i})+\beta\,\cos(\mathbf{g}_u,\mathbf{f}^{\mathrm{img}}_{i}).
\end{equation}

The top-\(K\) objects \(\mathcal{Q}_u\) are retained; each is expanded with its spatial neighbors via \(\mathcal{E}^{\mathrm{spa}}_t\) and with its history via \(\mathcal{E}^{\mathrm{temp}}\). The resulting task-oriented subgraph \(\mathcal{S}_u\) encapsulates the relevant objects, their relations, and temporal evolution. \(\mathcal{S}_u\) is then serialized into a lightweight JSON-style description that lists, for each node, its class \(y\), centroid \(\mathbf{c}\), size \(\mathbf{s}\), salient spatial relations, and motion history, along with a concise summary of recent scene dynamics. This description, paired with the instruction \(u\) and the corresponding RGB video, is passed to the LVLM planner \(\Psi_{\mathrm{plan}}\). The planner outputs a sequence of high-level actions \(\pi=(a_1,\dots,a_M)\) with grounded arguments (e.g., centroids and sizes), which are parsed into skill parameters for downstream controllers.

By conditioning on \(\mathcal{S}_u\) rather than the full graph, the planner focuses on instruction-relevant entities while preserving open-vocabulary flexibility. Crucially, because each frame-level graph \(\mathcal{M}_n\) stores both its capture timestamp \(\tau_n\) and estimated latency \(\Delta T_n\), the planner can retrieve the scene state aligned with the user’s instruction time \(\tau_u\). This mechanism explicitly grounds planning in the operator’s perspective on the local side, thereby mitigating errors introduced by transmission latency and enabling temporally consistent reasoning on the remote side.

\section{Experiments} \label{sec:exp}

\subsection{Dataset and Implementation Details}\label{subsec:dataset}
Our evaluation is conducted on a custom dataset designed to capture the challenges of dynamic, non-static tabletop environments. The dataset consists of 70 RGB-D video sequences recorded with an Intel RealSense D435i from a fixed camera viewpoint. The sequences span four everyday settings: office desk, lab workbench, storage shelf, and kitchen. Unlike static benchmarks, these videos feature continuous scene evolution, where objects are moved, occluded, rotated, duplicated, or removed. Each sequence is paired with a JSON annotation file containing multiple instruction–action tuples. Depth information complements RGB for accurate grounding of geometry, occlusion, and physical interactions. Instruction–action tuples were initially generated by a vision–language model and subsequently verified and refined by human annotators. The dataset is intended to evaluate the scene representation precision and planners that translate natural language into grounded, temporally coherent action sequences that remain robust under dynamic conditions, transmission delays, and multiple contingencies.

For implementation, the detection–description LVLM is instantiated with Qwen2.5-VL-7B \cite{bai2025qwen2}. Open-vocabulary object detection and segmentation are performed using Grounded-SAM2-Large \cite{sam2} in combination with GroundingDINO-Swin-B \cite{groundingdino}. For robot planning, the LVLM planner is implemented with Qwen2.5-VL-32B \cite{bai2025qwen2}.

\subsection{Static Representation Construction}\label{subsec:static}

To evaluate the quality of the proposed static scene representation, we conducted experiments on the Replica dataset\cite{replica1}, which provides high-fidelity indoor environments. Following the protocol of ConceptGraphs, we selected seven different scenes and, for each scene, randomly sampled static images from five distinct viewpoints to construct scene graphs.

Because our representation is designed for open-vocabulary settings, automated evaluation of nodes and edges is unreliable: object categories and relational boundaries under open vocabulary cannot be strictly compared with fixed labels or rules. We therefore adopted a human evaluation protocol. Three annotators independently judged the correctness of each node and edge, and an element was considered correct if at least two annotators agreed. This majority-vote strategy mitigates subjective bias and yields a stable estimate of accuracy.


\begin{table}[ht!]
\centering
\caption{Comparison between ConceptGraph and ST-OVSG.}\label{tab:compare}
\renewcommand{\arraystretch}{1.15}
\setlength{\tabcolsep}{5pt}
\begin{tabular}{l|cc|cc}
\hline
 & \multicolumn{2}{c|}{\textbf{ConceptGraph}} & \multicolumn{2}{c}{\textbf{ST-OVSG}} \\
\cline{2-5}
scene   & node prec. & edge prec.\footnotemark[1] & node prec. & edge prec.\footnotemark[2] \\
\hline
room0   & 0.78 & 0.91 & 0.69 & 0.94 \\
room1   & 0.77 & 0.93 & 0.85 & 0.95 \\
room2   & 0.66 & 1.00 & 0.70 & 0.53\\
office0 & 0.65 & 0.88 & 0.73 & 0.35\\
office1 & 0.65 & 0.90 & 0.67 & 0.72 \\
office2 & 0.75 & 0.82 & 0.76 & 0.73 \\
office3 & 0.68 & 0.79 & 0.72 & 0.47 \\
\hline
\rowcolor{gray!15}\textbf{total} & 0.71 & \textbf{0.88} & \textbf{0.74} & 0.67 \\
\hline
\end{tabular}
\end{table}

\footnotetext[1]{Edge precision corresponds to spatial edges in ConceptGraph.}
\footnotetext[2]{Edge precision corresponds to spatial edges in ST-OVSG.}

The results are shown in Table~\ref{tab:compare}. Our method achieved a node accuracy of 74\%, outperforming ConceptGraphs\cite{conceptgraphs}, while edge accuracy reached 67\%, slightly lower than ConceptGraphs. Node errors mainly stem from two sources: (1) the visual–language model occasionally introduced semantic bias, leading to missing or misclassified objects; and (2) the detection and segmentation modules struggled with small or visually subtle objects. The lower edge accuracy is largely explained by the single-view setting, which inherently limits relational coverage compared with multi-view methods, compounded by the restricted visibility of certain chosen viewpoints. Nevertheless, under favorable viewpoints, the relational modeling remains robust and yields accurate representations. These static results establish a baseline for subsequent experiments on dynamic environments, where temporal reasoning and latency-awareness play a central role.

\subsection{Dynamic Representation Construction}\label{subsec:dynamic}

ST-OVSG is designed to model dynamic scenes by capturing both spatial changes in the environment and temporal evolution of objects, enabling intent reasoning and key information retrieval under teleoperation. To evaluate the fidelity of the constructed representations, we tested on the dataset that is described in Subsection~\ref{subsec:dataset}. For each sequence, accuracies were measured for nodes, spatial edges, and temporal edges. As in the static case, open-vocabulary evaluation required human judgment: a node was correct if its localization and caption were semantically accurate, a spatial edge was correct if both endpoints and their relation label matched the ground truth, and a temporal edge was correct if cross-frame association tracked the same physical object with the correct change label.

Modeling dynamic videos is substantially more challenging than static images, which explains the lower accuracies observed. Motion blur, viewpoint shifts, and occlusions destabilize open-vocabulary detections. Objects may enter or leave the field of view or appear in unusual poses, which complicates cross-frame association. Near-duplicate instances further stress relation inference and long-range identity tracking. These factors accumulate across time, so even small frame-level errors can cascade into graph-level inconsistencies.

\begin{table}[ht!]
\centering
\caption{Evaluation of node, spatial-edge, and temporal-edge accuracy across dynamic video scenes.}\label{tab:video}
\renewcommand{\arraystretch}{1.12}
\setlength{\tabcolsep}{8pt}
\begin{tabular}{l|c|c|c}
\hline
scene & node & spatial edge & temporal edge \\
\hline
office desk     & 0.66 & 0.65 & 0.61 \\
lab workbench   & 0.76 & 0.67 & 0.43 \\
storage shelf   & 0.68 & 0.56 & 0.51 \\
kitchen         & 0.57 & 0.51 & 0.65 \\
\hline
\rowcolor{gray!15}\textbf{total} & \textbf{0.66} & \textbf{0.64} & \textbf{0.61} \\
\hline
\end{tabular}
\end{table}

Results are summarized in Table~\ref{tab:video}, with accuracies of \(A_{\mathrm{node}}=\text{66}\%\), \(A_{\mathrm{spa}}=\text{64}\%\), and \(A_{\mathrm{tmp}}=\text{61}\%\). Most node and spatial-edge errors arise from motion blur or atypical poses that shift visual features and mislead the LVLM detector. Temporal-edge errors are dominated by short-term occlusions, pose changes, and the introduction of visually similar distractors. Compared with static scenes (Subsection.~\ref{subsec:static}), the overall accuracy decreases, but ST-OVSG still consistently maintains temporal identity and evolving relations. This capability is particularly valuable for teleoperation, where transmission latency requires reasoning over past scene states rather than the delayed present.

\subsection{Latency-Robustness Experiment} \label{subsec:latency}

The second set of experiments evaluates whether the LVLM planner can remain \emph{latency-aware} by exploiting the per-frame timestamp and latency tags \((\tau_t,\lambda_t)\) stored in ST-OVSG. These tags enable the planner to align grounding with the scene state that existed when the operator issued the instruction, rather than the delayed state available at the robot. This setting reflects the teleoperation scenario, where video streams are transmitted from the remote robot to the local operator, and natural-language commands are then sent back to the remote robot and its LVLM planner, introducing bidirectional communication delays.

We designed tasks in which latency fundamentally changes the grounding: (i) \textbf{Occlusion-after-command}: the target is visible at issue time but becomes occluded before robot received command; (ii) \textbf{Target moved}: the target is displaced after issue time; (iii) \textbf{Same-class distractor}: a visually similar object is introduced after issue time; (iv) \textbf{Moved reference}: the instruction refers to an object relative to a landmark (e.g., “the fruit next to the phone”), and the landmark is moved after issue time. These scenarios are intentionally adversarial for non–latency-aware planners, which only operate on the most recent frame without historical alignment.

Artificial delays in the range \(0.25\mathrm{s}\)–\(5\mathrm{s}\) were injected between the local operator and the remote robot+planner. For each trial, the issue timestamp was recorded, and the planner received the corresponding task-oriented subgraph together with the RGB video stream. A trial was counted as successful only if (a) the planner selected the correct target object rather than a distractor, and (b) the grounded 3D parameters (e.g., centroid and size) were sufficiently accurate for execution. Across 17 trials, ST-OVSG achieved a success rate of \(70.5\%\).

Failure cases were dominated by residual identity switches under long occlusions, missed detections of small or subtle objects, and unstable temporal associations caused by motion blur or unusual poses. These are consistent with the challenges observed in dynamic-scene modeling (Subsection.~ \ref{subsec:dynamic}). Nonetheless, the results demonstrate that latency tags allow instructions to be grounded to the correct historical state, substantially reducing referential errors caused by delayed perception and communication. This provides concrete evidence that ST-OVSG enhances robustness in teleoperation settings where transmission latency is unavoidable.

\begin{figure*} [ht!]
     \centering
     \includegraphics[width=\textwidth]{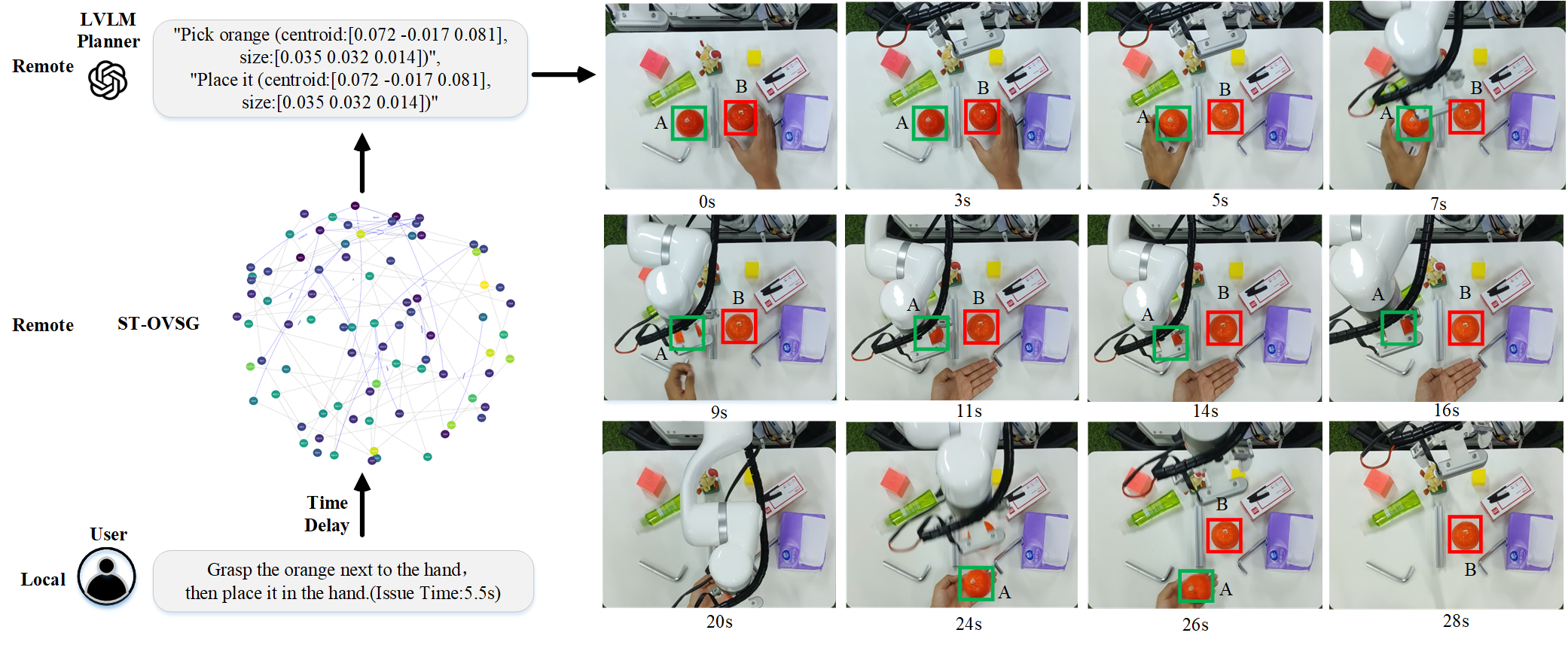}
     \caption{Execution process of the proposed method in a task. Left: users provide a natural-language grasp-and-place instruction at the local side (issue at 5.5s and communication latency is 500ms). ST-OVSG builds a time-aware, open-vocabulary scene graph, and based on this, the LVLM generates a latency-aware grasp-and-place plan at the remote side. Right: the robot executes the plan in sequence: it approaches the target, performs a stable grasp, transports the object smoothly, and places it safely at the designated location.}
     \label{fig3}
     \vspace{-1.5em}
\end{figure*}

\subsection{Ablation under Zero-Latency Planning}

This ablation study isolates the contribution of ST-OVSG to planning quality under the idealized condition of zero communication delay. The same LVLM planner (Qwen2.5-VL-32B) is used in all conditions. We compare: (i) \emph{with ST-OVSG}, in which the planner receives the task-conditioned JSON subgraph \(\mathcal{S}_u\) in addition to the instruction and RGB frames; (ii) \emph{without ST-OVSG}, in which only the instruction and frames are provided; and (iii) the \emph{ground-truth (GT) plan}, which serves as the reference. 

For evaluation, we use the sentence-transformers model \cite{reimers-2019-sentence-bert} to compute embedding-based similarity between predicted plans and GT plans. Each action sequence is linearized into imperative sentences, normalized, and encoded, after which cosine similarity is computed at the sequence level. This metric is tolerant to paraphrasing and lexical variation, which are common in open-vocabulary outputs, though it cannot fully bridge semantic mismatches between equivalent but differently worded plans. With ST-OVSG, the average similarity score is 0.1702, compared to 0.164 without ST-OVSG. While the absolute improvement is small, this result reflects a consistent trend: adding structured scene information via ST-OVSG does not degrade planning quality, and in some cases marginally improves alignment to the reference. The overall scores are relatively low, which we attribute to the intrinsic difficulty of evaluating open-vocabulary plans against GT text. In practice, many predicted actions were semantically correct but expressed with different phrasing or level of detail, which lowers embedding-based similarity without indicating execution failure. 

This experiment therefore suggests that ST-OVSG provides auxiliary cues that can stabilize plan generation, though its benefit is more pronounced in the delayed, dynamic scenarios (Sec.~\ref{subsec:latency}) where historical alignment is critical.

\subsection{Real Robot Experiment}
To validate the effectiveness of the proposed method, we conduct a teleoperation grasp-and-place experiment with a local human operator and a remote xArm 6 equipped with a RealSense D435i camera\cite{d435i}. As shown in Fig. \ref{fig3}, the hand is initially positioned near Orange B, moves toward Orange A within a short time window, and then relocates to a designated position. When the local feedback interface indicates that the hand is approaching Orange A, the operator issues the command: “Grasp the orange next to the hand, then place it in the hand.” Due to a 500ms communication delay, the command reaches the remote at 6.0s, at which point the robotic arm begins executing the corresponding plan. Compared to methods that ignore latency (often manifesting as misgrabbing Orange B, oscillating between the two oranges, or passive stagnation), after brief evaluation or fine-tuning, this system enables the end-effector trajectory of the robotic arm to monotonically converge toward Orange A without oscillation or path crossing. During the approach, the gripper's pose remains well-aligned with the pose of Orange A, ensuring a successful grasp. The orange is then transported along a smooth trajectory and is placed securely in the hand. In summary, the method maintains temporally consistent alignment with the target object even in the presence of execution delays.
\section{Discussion and Conclusion}
We proposed \textbf{ST-OVSG}, an open-vocabulary spatio-temporal scene graph that provides dynamic, latency-tagged, and retrieval-friendly environment representations to support teleoperation. Our method mitigates their impact by reducing the risk of operator instructions being misunderstood due to local–remote state mismatches. ST-OVSG progressively models dynamic environments into a task oriented spatio-temporal graph, highlighting task-relevant information, alleviating redundancy, and providing explicit spatial and temporal relations augmented with lightweight latency annotations. These properties enable consistent object tracking and retrieval across time that prior static representations lack but are critical for reliable teleoperation. Experiments demonstrate that, when coupled with an LVLM planner, ST-OVSG enhances temporal reasoning and supports delay-aware decision making without task-specific training, thereby reducing referential errors under delayed perception and communication. In future work, we plan to strengthen robustness under challenging visual conditions and pursue end-to-end integration of scene modeling, delay annotation, and planning to further improve long-horizon teleoperation performance.

\bibliographystyle{IEEEtran}
\bibliography{root}

\begin{thebibliography}{10}
\providecommand{\url}[1]{#1}
\csname url@samestyle\endcsname
\providecommand{\newblock}{\relax}
\providecommand{\bibinfo}[2]{#2}
\providecommand{\BIBentrySTDinterwordspacing}{\spaceskip=0pt\relax}
\providecommand{\BIBentryALTinterwordstretchfactor}{4}
\providecommand{\BIBentryALTinterwordspacing}{\spaceskip=\fontdimen2\font plus
\BIBentryALTinterwordstretchfactor\fontdimen3\font minus \fontdimen4\font\relax}
\providecommand{\BIBforeignlanguage}[2]{{%
\expandafter\ifx\csname l@#1\endcsname\relax
\typeout{** WARNING: IEEEtran.bst: No hyphenation pattern has been}%
\typeout{** loaded for the language `#1'. Using the pattern for}%
\typeout{** the default language instead.}%
\else
\language=\csname l@#1\endcsname
\fi
#2}}
\providecommand{\BIBdecl}{\relax}
\BIBdecl

\bibitem{intro1}
S.~Avgousti, E.~G. Christoforou, A.~S. Panayides, S.~Voskarides, C.~Novales, L.~Nouaille, C.~S. Pattichis, and P.~Vieyres, ``Medical telerobotic systems: current status and future trends,'' \emph{Biomedical engineering online}, vol.~15, no.~1, p.~96, 2016.

\bibitem{intro2}
H.~Chitikena, F.~Sanfilippo, and S.~Ma, ``Robotics in search and rescue (sar) operations: An ethical and design perspective framework for response phase,'' \emph{Applied Sciences}, vol.~13, no.~3, p. 1800, 2023.

\bibitem{intro3}
H.~Zhou, M.~Ding, W.~Peng, M.~Tomizuka, L.~Shao, and C.~Gan, ``Generalizable long-horizon manipulations with large language models,'' \emph{arXiv preprint arXiv:2310.02264}, 2023.

\bibitem{intro4}
S.~B. Kamtam, Q.~Lu, F.~Bouali, O.~C. Haas, and S.~Birrell, ``Network latency in teleoperation of connected and autonomous vehicles: A review of trends, challenges, and mitigation strategies,'' \emph{Sensors (Basel, Switzerland)}, vol.~24, no.~12, p. 3957, 2024.

\bibitem{nlmap}
B.~Chen, F.~Xia, B.~Ichter, K.~Rao, K.~Gopalakrishnan, M.~S. Ryoo, A.~Stone, and D.~Kappler, ``Open-vocabulary queryable scene representations for real world planning,'' \emph{arXiv preprint arXiv:2209.09874}, 2022.

\bibitem{vlp}
C.~Huang, O.~Mees, A.~Zeng, and W.~Burgard, ``Visual language maps for robot navigation,'' \emph{arXiv preprint arXiv:2210.05714}, 2022.

\bibitem{conceptgraphs}
Q.~Gu, A.~Kuwajerwala, S.~Morin, K.~M. Jatavallabhula, B.~Sen, A.~Agarwal, C.~Rivera, W.~Paul, K.~Ellis, R.~Chellappa \emph{et~al.}, ``Conceptgraphs: Open-vocabulary 3d scene graphs for perception and planning,'' in \emph{2024 IEEE International Conference on Robotics and Automation (ICRA)}.\hskip 1em plus 0.5em minus 0.4em\relax IEEE, 2024, pp. 5021--5028.

\bibitem{Hungarian1}
J.~Munkres, ``Algorithms for the assignment and transportation problems,'' \emph{Journal of the society for industrial and applied mathematics}, vol.~5, no.~1, pp. 32--38, 1957.

\bibitem{Hungarian2}
H.~W. Kuhn, ``The hungarian method for the assignment problem,'' \emph{Naval research logistics quarterly}, vol.~2, no. 1-2, pp. 83--97, 1955.

\bibitem{brown}
T.~Brown, B.~Mann, N.~Ryder, M.~Subbiah, J.~D. Kaplan, P.~Dhariwal, A.~Neelakantan, P.~Shyam, G.~Sastry, A.~Askell \emph{et~al.}, ``Language models are few-shot learners,'' \emph{Advances in neural information processing systems}, vol.~33, pp. 1877--1901, 2020.

\bibitem{liu2023reflect}
Z.~Liu, A.~Bahety, and S.~Song, ``Reflect: Summarizing robot experiences for failure explanation and correction,'' \emph{arXiv preprint arXiv:2306.15724}, 2023.

\bibitem{saycan}
M.~Ahn, A.~Brohan, N.~Brown, Y.~Chebotar, O.~Cortes, B.~David, C.~Finn, C.~Fu, K.~Gopalakrishnan, K.~Hausman \emph{et~al.}, ``Do as i can, not as i say: Grounding language in robotic affordances,'' \emph{arXiv preprint arXiv:2204.01691}, 2022.

\bibitem{PaLM-E}
D.~Driess, F.~Xia, M.~S. Sajjadi, C.~Lynch, A.~Chowdhery, A.~Wahid, J.~Tompson, Q.~Vuong, T.~Yu, W.~Huang \emph{et~al.}, ``Palm-e: An embodied multimodal language model,'' 2023.

\bibitem{progprompt}
I.~Singh, V.~Blukis, A.~Mousavian, A.~Goyal, D.~Xu, J.~Tremblay, D.~Fox, J.~Thomason, and A.~Garg, ``Progprompt: Generating situated robot task plans using large language models,'' \emph{arXiv preprint arXiv:2209.11302}, 2022.

\bibitem{slam1}
J.~Civera, D.~G{\'a}lvez-L{\'o}pez, L.~Riazuelo, J.~D. Tard{\'o}s, and J.~M.~M. Montiel, ``Towards semantic slam using a monocular camera,'' in \emph{2011 IEEE/RSJ international conference on intelligent robots and systems}.\hskip 1em plus 0.5em minus 0.4em\relax IEEE, 2011, pp. 1277--1284.

\bibitem{slam2}
L.~Zhang, L.~Wei, P.~Shen, W.~Wei, G.~Zhu, and J.~Song, ``Semantic slam based on object detection and improved octomap,'' \emph{IEEE Access}, vol.~6, pp. 75\,545--75\,559, 2018.

\bibitem{slam3}
Z.~Zhu, S.~Peng, V.~Larsson, W.~Xu, H.~Bao, Z.~Cui, M.~R. Oswald, and M.~Pollefeys, ``Nice-slam: Neural implicit scalable encoding for slam,'' in \emph{Proceedings of the IEEE/CVF conference on computer vision and pattern recognition}, 2022, pp. 12\,786--12\,796.

\bibitem{slam4}
E.~Sucar, S.~Liu, J.~Ortiz, and A.~J. Davison, ``imap: Implicit mapping and positioning in real-time,'' in \emph{Proceedings of the IEEE/CVF international conference on computer vision}, 2021, pp. 6229--6238.

\bibitem{detection1}
M.~Runz, M.~Buffier, and L.~Agapito, ``Maskfusion: Real-time recognition, tracking and reconstruction of multiple moving objects,'' in \emph{2018 IEEE international symposium on mixed and augmented reality (ISMAR)}.\hskip 1em plus 0.5em minus 0.4em\relax IEEE, 2018, pp. 10--20.

\bibitem{detection2}
J.~McCormac, R.~Clark, M.~Bloesch, A.~Davison, and S.~Leutenegger, ``Fusion++: Volumetric object-level slam,'' in \emph{2018 international conference on 3D vision (3DV)}.\hskip 1em plus 0.5em minus 0.4em\relax IEEE, 2018, pp. 32--41.

\bibitem{detection3}
J.~Qian, V.~Chatrath, J.~Yang, J.~Servos, A.~P. Schoellig, and S.~L. Waslander, ``Pocd: Probabilistic object-level change detection and volumetric mapping in semi-static scenes,'' \emph{arXiv preprint arXiv:2205.01202}, 2022.

\bibitem{detection4}
K.~Li, D.~DeTone, Y.~F.~S. Chen, M.~Vo, I.~Reid, H.~Rezatofighi, C.~Sweeney, J.~Straub, and R.~Newcombe, ``Odam: Object detection, association, and mapping using posed rgb video,'' in \emph{Proceedings of the IEEE/CVF International Conference on Computer Vision}, 2021, pp. 5998--6008.

\bibitem{boundingbox1}
C.~R. Qi, X.~Chen, O.~Litany, and L.~J. Guibas, ``Imvotenet: Boosting 3d object detection in point clouds with image votes,'' in \emph{Proceedings of the IEEE/CVF conference on computer vision and pattern recognition}, 2020, pp. 4404--4413.

\bibitem{boundingbox2}
D.~Xu, D.~Anguelov, and A.~Jain, ``Pointfusion: Deep sensor fusion for 3d bounding box estimation,'' in \emph{Proceedings of the IEEE conference on computer vision and pattern recognition}, 2018, pp. 244--253.

\bibitem{scene_graph1}
N.~Hughes, Y.~Chang, and L.~Carlone, ``Hydra: A real-time spatial perception system for 3d scene graph construction and optimization,'' \emph{arXiv preprint arXiv:2201.13360}, 2022.

\bibitem{scene_graph2}
I.~Armeni, Z.-Y. He, J.~Gwak, A.~R. Zamir, M.~Fischer, J.~Malik, and S.~Savarese, ``3d scene graph: A structure for unified semantics, 3d space, and camera,'' in \emph{Proceedings of the IEEE/CVF international conference on computer vision}, 2019, pp. 5664--5673.

\bibitem{clip}
A.~Radford, J.~W. Kim, C.~Hallacy, A.~Ramesh, G.~Goh, S.~Agarwal, G.~Sastry, A.~Askell, P.~Mishkin, J.~Clark \emph{et~al.}, ``Learning transferable visual models from natural language supervision,'' in \emph{International conference on machine learning}.\hskip 1em plus 0.5em minus 0.4em\relax PmLR, 2021, pp. 8748--8763.

\bibitem{vild}
X.~Gu, T.-Y. Lin, W.~Kuo, and Y.~Cui, ``Open-vocabulary object detection via vision and language knowledge distillation,'' \emph{arXiv preprint arXiv:2104.13921}, 2021.

\bibitem{bai2025qwen2}
S.~Bai, K.~Chen, X.~Liu, J.~Wang, W.~Ge, S.~Song, K.~Dang, P.~Wang, S.~Wang, J.~Tang \emph{et~al.}, ``Qwen2. 5-vl technical report,'' \emph{arXiv preprint arXiv:2502.13923}, 2025.

\bibitem{sam2}
N.~Ravi, V.~Gabeur, Y.-T. Hu, R.~Hu, C.~Ryali, T.~Ma, H.~Khedr, R.~R{\"a}dle, C.~Rolland, L.~Gustafson \emph{et~al.}, ``Sam 2: Segment anything in images and videos,'' \emph{arXiv preprint arXiv:2408.00714}, 2024.

\bibitem{groundingdino}
S.~Liu, Z.~Zeng, T.~Ren, F.~Li, H.~Zhang, J.~Yang, Q.~Jiang, C.~Li, J.~Yang, H.~Su \emph{et~al.}, ``Grounding dino: Marrying dino with grounded pre-training for open-set object detection,'' in \emph{European conference on computer vision}.\hskip 1em plus 0.5em minus 0.4em\relax Springer, 2024, pp. 38--55.

\bibitem{replica1}
J.~Straub, T.~Whelan, L.~Ma, Y.~Chen, E.~Wijmans, S.~Green, J.~J. Engel, R.~Mur-Artal, C.~Ren, S.~Verma, A.~Clarkson, M.~Yan, B.~Budge, Y.~Yan, X.~Pan, J.~Yon, Y.~Zou, K.~Leon, N.~Carter, J.~Briales, T.~Gillingham, E.~Mueggler, L.~Pesqueira, M.~Savva, D.~Batra, H.~M. Strasdat, R.~D. Nardi, M.~Goesele, S.~Lovegrove, and R.~Newcombe, ``The {R}eplica dataset: A digital replica of indoor spaces,'' \emph{arXiv preprint arXiv:1906.05797}, 2019.

\bibitem{reimers-2019-sentence-bert}
\BIBentryALTinterwordspacing
N.~Reimers and I.~Gurevych, ``Sentence-bert: Sentence embeddings using siamese bert-networks,'' in \emph{Proceedings of the 2019 Conference on Empirical Methods in Natural Language Processing}.\hskip 1em plus 0.5em minus 0.4em\relax Association for Computational Linguistics, 11 2019. [Online]. Available: \url{https://arxiv.org/abs/1908.10084}
\BIBentrySTDinterwordspacing

\bibitem{d435i}
L.~Keselman, J.~Iselin~Woodfill, A.~Grunnet-Jepsen, and A.~Bhowmik, ``Intel realsense stereoscopic depth cameras,'' in \emph{Proceedings of the IEEE conference on computer vision and pattern recognition workshops}, 2017, pp. 1--10.

\end{thebibliography}

\end{document}